\documentclass{article}
\usepackage{acra}
\usepackage{amsmath,amsfonts}
\usepackage{graphicx}
\usepackage{booktabs}
\usepackage{colortbl}
\definecolor{rowsepgrey}{gray}{0.65}
\newcommand{\rowsep}{\arrayrulecolor{rowsepgrey}\specialrule{0.4pt}{3pt}{3pt}\arrayrulecolor{black}}  
\usepackage{multirow}
\usepackage{array}
\usepackage{tabularx}
\usepackage{url}
\usepackage[font=small]{caption}
\usepackage{placeins}

\title{On Shaky Ground: Characterising Non-Rigid Base Effects \\in Mobile Manipulation}
\author{Zongyuan Zhang$^{1,2,3,*}$, Christopher Lehnert$^{1,2,3}$, \\ {\bf Will N. Browne$^{1,2,3}$, Jonathan M. Roberts$^{1,2,3}$} \\
{\rm $^{1}$Faculty of Engineering, $^{2}$Australian Cobotics Centre, $^{3}$QUT Centre for Robotics} \\
{\rm Queensland University of Technology, 2 George St, Brisbane, QLD 4000, Australia} \\
{\rm $^{*}$Corresponding author: \texttt{z203.zhang@hdr.qut.edu.au}}}

\begin{document}

\maketitle

\begin{abstract}
Mobile robot manipulators extend contact-task capability beyond the workspace of fixed robot installations, but their bases must also sustain the forces generated by the task. On lightweight platforms, base motion and support compliance can change the force and motion realised at the workpiece. How these non-rigid base effects manifest during sustained contact remains insufficiently characterised. This paper investigates non-rigid base effects in mobile robot manipulators through a sanding case study combining exploratory simulation with experiments on two lightweight, unactuated wheeled testbeds. Joint measurements of contact force, tool motion, and base vibration show that sustained contact coexisted with substantial task error: under low base--ground resistance, only 54--70\% of a 100~N command was realised, with tool-centre-point errors reaching 100~mm. Increasing base--ground resistance during contact improved force realisation and arrested further tool displacement, while base vibration did not vary consistently with force-tracking error in the tested conditions. These observations illustrate how non-rigid base behaviour can affect contact-task execution and highlight the value of examining force, motion, and vibration together when characterising lightweight mobile manipulation.
\end{abstract}

\begin{figure}[t]
\centering
\includegraphics[width=\columnwidth]{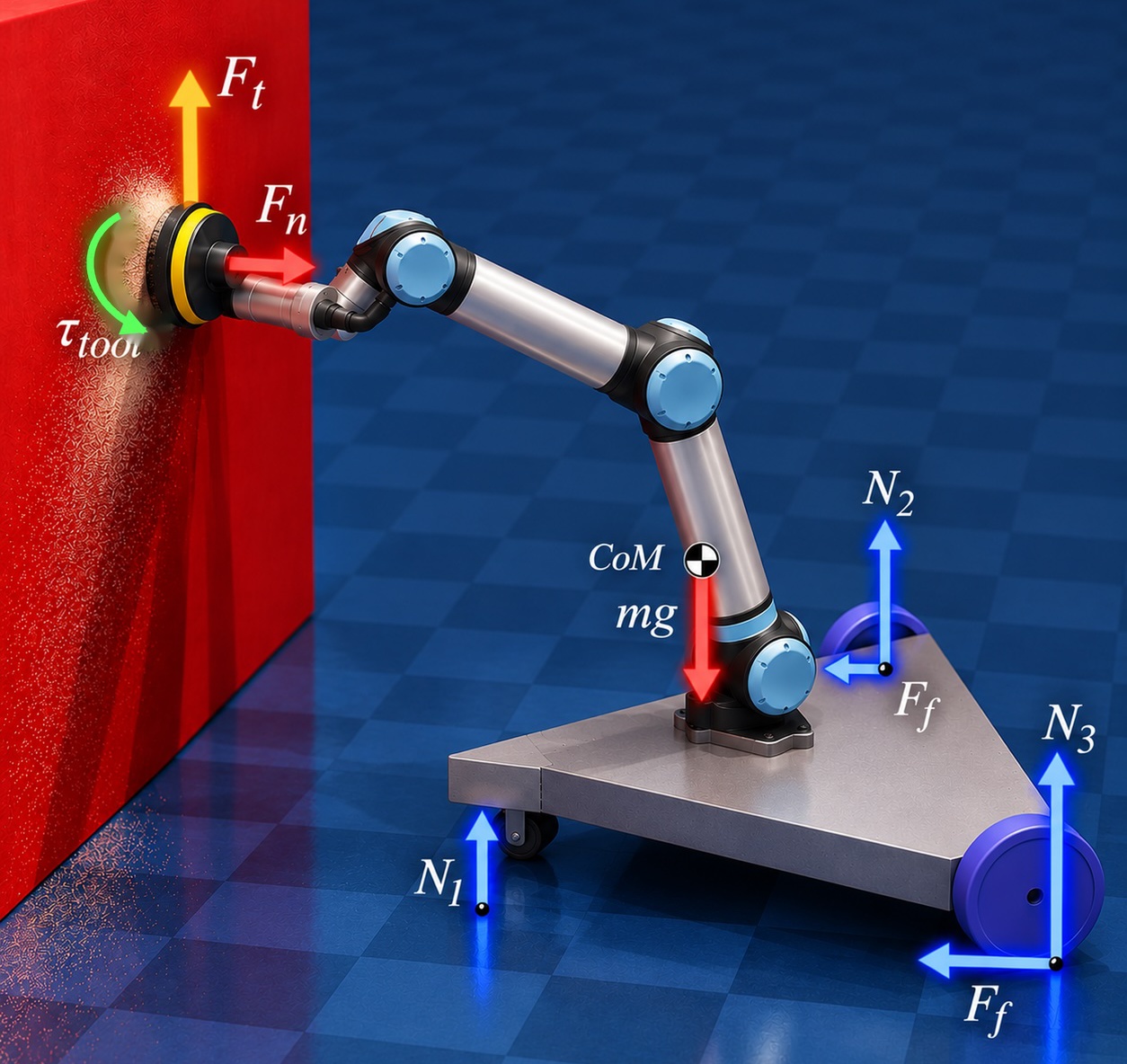}
\caption{Contact forces acting on a mobile manipulator during sanding. The contact forces $F_n$ and $F_t$ and the tool torque $\tau_{tool}$ are transmitted through the manipulator to the base and balanced by the ground support forces $N_1$--$N_3$ and the friction forces $F_f$, with the weight $mg$ acting at the centre of mass (CoM). Limited traction and compliance at the base--ground interface can lead to base motion, loss of stability, and vibration.}
\label{fig:free_body}
\end{figure}

\section{Introduction}
Fixed-base industrial manipulators are widely used for contact-rich processes such as sanding, polishing, and assembly. Their rigid bases enable accurate motion and force control, but limit operation to a fixed workspace. Mounting the manipulator on a mobile robot extends the operating workspace and allows the tool to be carried between work areas, making mobile manipulators attractive for tasks involving large or distributed structures such as aircraft, ships, and building surfaces \cite{mobile_survey}.

This mobility, however, introduces additional challenges in contact tasks. Contact forces are transmitted through the manipulator and the mobile base to the ground. Existing mobile machining systems often use deliberately heavy automated guided vehicles or track-mounted platforms \cite{mobile_heavy,mobile_AGV1,mobile_AGV2,train_sanding_yang} to increase stiffness and stability during contact. Lightweight mobile platforms are easier to move and deploy, but their lower mass, limited wheel traction, and compliance at the wheel--ground interface can lead to base motion and vibration under contact load.

Figure~\ref{fig:free_body} illustrates these effects. Normal and tangential forces at the sanding tool are transmitted through the manipulator to the base. If the horizontal reaction exceeds the available traction, the base can slip. If the resultant wrench moves the zero-moment point outside the support polygon, the system can lose static stability. Compliance in the tyres, suspension, or ground surface can also introduce oscillation of the base and manipulator. These behaviours are collectively referred to as \emph{non-rigid base effects}. They affect contact-task performance because base motion contributes to tool centre point (TCP) error \cite{mobile_acc}, while base vibration and deformation can disturb the contact force. Under force control, the arm commands tool advance relative to its base, whereas the contact force depends on the tool--workpiece compression. Base motion away from the workpiece under the contact reaction can therefore absorb part of the commanded advance and contribute to a sustained force shortfall and to undesired tool motion relative to the workpiece, even while contact is maintained.

Existing work has examined mobile-manipulator stability, wheel slip, machining accuracy, compliance, and contact-force control, but there is limited experimental characterisation of how base motion manifests simultaneously in realised contact force and tool motion relative to the workpiece during sustained contact on lightweight mobile manipulators. This paper addresses this gap through a sanding case study combining exploratory simulation with experiments on two instrumented lightweight wheeled testbeds. Their unactuated bases serve as mechanical surrogates for mobile robot bases (Section~\ref{sec:platforms}). The study addresses three questions: (i) What non-rigid base effects arise under contact forces? (ii) How do base--environment coupling, arm posture, tool excitation, and support compliance affect force realisation, tool motion, and base vibration? (iii) Do the principal response patterns persist across different lightweight platforms?

The contributions are:
\begin{itemize}
    \item An experimental characterisation of non-rigid base effects in force-controlled sanding, showing that sustained contact coexisted with force shortfall and TCP error, and that increasing base--ground resistance during contact improved force realisation and arrested further tool displacement;
    \item An assessment of how these effects vary with commanded force, arm posture, tool excitation, and support compliance on two lightweight mobile platforms, with base-vibration measurements providing a complementary view of the dynamic response.
\end{itemize}
These measurements identify force and motion responses that future active compensation strategies on mobile robots would need to address.

\section{Related Work}

\subsection{Robotic Surface Finishing}

Robotic grinding and sanding require simultaneous path tracking and regulation of the tool--workpiece interaction. Process studies relate abrasive properties, tool speed, feed rate, and normal force to material removal and surface quality \cite{grinding_model,grinding_factor2,MRR_Zhang,MRR_Li_model}. Reviews of mobile robotic machining and grinding identify limited structural stiffness, motion accuracy, and process vibration as persistent limitations \cite{mobile_survey,Li2026_mobile_grinding_review}.

Previous studies have addressed grinding vibration through feed-speed and force control, and through posture selection \cite{Vibration_control2}, force-controlled end-effectors for flexible workpieces \cite{dampening_endeffector1}, and tool-holder compliance design \cite{dampening_endeffector3}. These approaches mainly address vibration and compliance associated with the manipulator, tool, or workpiece, while the influence of a mobile base and its interaction with the ground has received less attention.

\subsection{Mobile Machining and Base Stability}

Mobile robotic machining extends the workspace available for large aircraft, ships, rail vehicles, and other large-scale structures \cite{mobile_survey,Li2026_mobile_grinding_review}. Existing systems commonly use heavy automated guided vehicles, rail-mounted bases, external metrology, or compliant grinding heads to recover machining accuracy \cite{mobile_heavy,mobile_acc,mobile_AGV2,train_sanding_yang}. These systems typically rely either on heavy mobile platforms or on specially designed, structured environments to provide sufficient stability, reducing the mobility and ease of deployment that motivate lightweight mobile manipulators.

The tip-over stability of mobile manipulators has a substantial theoretical foundation. Papadopoulos and Rey \shortcite{Papadopoulos1996_tipover} introduced the force--angle stability margin for systems subject to external and inertial loads. Garnier \emph{et al.}\ \shortcite{mobile_ship1} analysed the dynamic tip-over stability of a mobile manipulator for naval-manufacturing tasks and showed that joint speed, joint acceleration, and structure height affect its stability. More recent work derives individual wheel support forces for direct monitoring of contact loss \cite{Petrovic2022_supporting_forces}, while experiments with a UR5 on a mobile base show that high, cantilevered payloads can destabilise the platform during navigation \cite{Bostelman2017_stability}. Hence, force effects on the mobile base are worth investigating.

Wheel slip is often excluded by the pure-rolling assumption. Galicki \shortcite{Galicki2023_slip} treated wheel-slip reaction forces and unstructured end-effector forces as disturbances in finite-time task-space control of mobile manipulators. Such slip must therefore be accounted for when lightweight robot platforms are adopted for contact tasks.

\subsection{Base Compliance and Physical Interaction}

Wang and Carmichael \shortcite{Flexible_Experiment_Wang} experimentally varied mobile-base rigidity during physical human--robot collaboration and found that base compliance affects task accuracy and human interaction. Their results show that the mechanical properties of the mobile base and its contact with the ground can directly influence physical interaction.

Gao \emph{et al.}\ \shortcite{Gao2026_motion_force} investigated dynamic base--arm coupling in mobile robot--environment interaction and developed a motion/force controller to compensate for the resulting disturbances. Their wall-cleaning experiments considered force-controlled interaction while the mobile base remained in motion.

Together, these studies show that base compliance and base motion can significantly affect physical interaction. However, experimental characterisation of how base motion manifests jointly in realised contact force and tool motion relative to the workpiece remains limited for sustained force-controlled contact on lightweight platforms intended to remain stationary. The present study examines this setting by measuring contact force, tool motion, and base vibration together.

\section{Exploratory Simulation}
\label{sec:simulation}

The simulation is used as an exploratory tool to identify potential failure modes, examine how they vary with different system and task parameters, and inform the physical experiments (see Section~\ref{sec:experiments}).

\subsection{Environment and Control}

\begin{figure*}[tp]
    \centering
    \begin{tabular}[b]{@{}c@{}}\includegraphics[height=3.4cm]{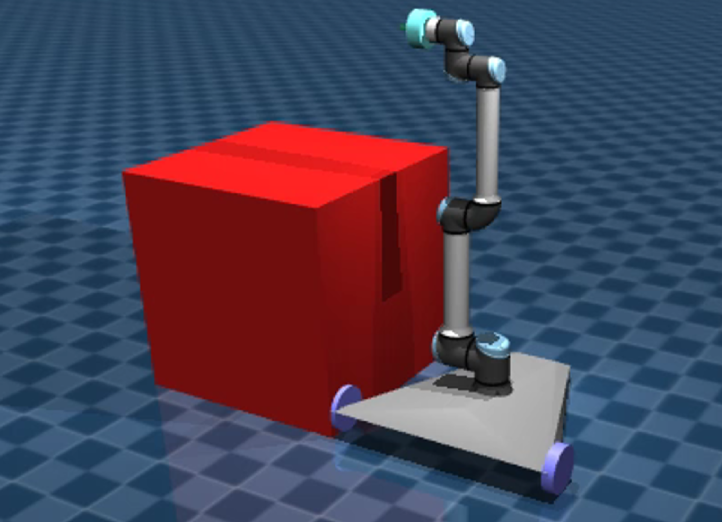}\\[1pt]\small (a) Robot and workpiece\end{tabular}\hfill
    \begin{tabular}[b]{@{}c@{}}\includegraphics[height=3.4cm]{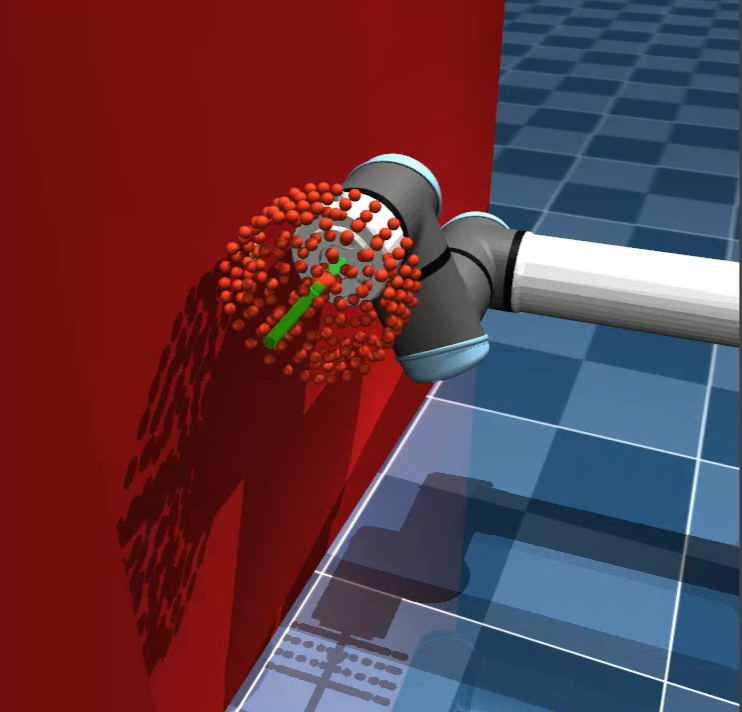}\\[1pt]\small (b) Compliant sanding tool\end{tabular}\hfill
    \begin{tabular}[b]{@{}c@{}}\includegraphics[height=3.4cm]{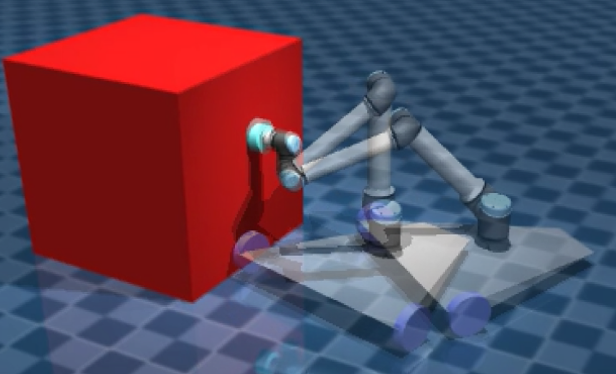}\\[1pt]\small (c) Base slip\end{tabular}\hfill
    \begin{tabular}[b]{@{}c@{}}\includegraphics[height=3.4cm]{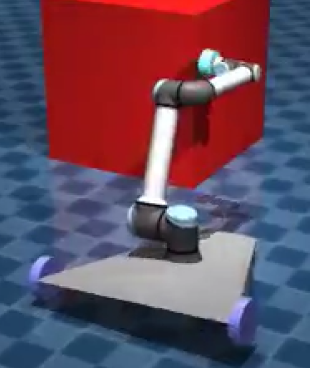}\\[1pt]\small (d) Loss of stability\end{tabular}
    \caption{Simulation environment and representative non-rigid base effects.}
    \label{fig:simulation_overview}
\end{figure*}

The simulation uses MuJoCo \cite{mujoco} with a UR10 manipulator on a differential-drive base facing a rigid vertical workpiece, as shown in Figure~\ref{fig:simulation_overview}a. The wheels and base structure are modelled as rigid, so the simulation captures traction and stability limits but does not include compliance at the wheel--ground contacts.

The end-effector carries a rotating sanding tool with a compliant pad and a virtual force/torque sensor. The pad is modelled using multiple MuJoCo contact points with compliant contact parameters, providing an approximate distributed contact over the sanding surface. A hybrid force--position controller regulates the normal contact force while controlling tangential motion \cite{Force_control_book_Siciliano}. Contact force, wheel friction, arm posture, mass distribution, and tool excitation are varied to examine the resulting failure modes.

\subsection{Observed Failure Modes}
\label{sec:sim_modes}

Heuristic parameter sweeps produced three repeatable behaviours (Figure~\ref{fig:simulation_overview}). First, contact excitation produced oscillation of the tool and base around the target pose. The oscillation increased with tool--workpiece friction, commanded normal force, and arm extension, while tool speed changed the dominant excitation frequency. These observations led to the use of separate force-variation and base-vibration metrics in the physical experiments.

Second, base slip occurred when the tangential forces required at the wheel--ground contacts exceeded the available traction. Slip became more likely with larger contact loads, lower wheel--ground friction, and postures that reduced the load on the driven wheels. This also motivated measuring tool motion and base response separately, since tool-path error alone cannot distinguish base motion from manipulator tracking error.

Third, the simulated robot lost stability when the combined gravitational and contact wrench moved the zero-moment point outside the support polygon. Extended postures and unfavourable mass distributions reduced the stability margin, identifying loss of stability as another potential failure mode under contact load.

These observations are qualitative. The simulated contact parameters were not identified from the physical platforms, and the rigid-wheel model does not represent compliance in the tyres, platform, or supporting surface. Such model mismatch is a source of the simulation-to-reality gap discussed in robotics \cite{Zhao2020_sim_to_real}. The simulation is therefore used to identify plausible failure modes and qualitative trends rather than to predict the hardware results quantitatively.

\section{Physical Experiments}
\label{sec:experiments}

The physical study uses two lightweight platforms, the \emph{cart} and the \emph{trolley} (Table~\ref{tab:platforms}), in three experiments on the same force-controlled sanding task (Section~\ref{sec:trials}). The experiments vary resistance to base motion and support compliance to characterise the resulting contact response. The two constructions provide contrasting support conditions for examining whether the principal effects persist across platforms.

Neither platform is an actuated mobile robot. They are used as surrogates for a lightweight mobile-robot base because they provide variable manipulator attachment points, are low cost, and allow repeatable base--ground conditions.

\begin{table}[t]
    \centering
    \footnotesize
    \renewcommand{\arraystretch}{1.15}
    \begin{tabular*}{\columnwidth}{@{\extracolsep{\fill}}llll@{}}
        \toprule
        & Simulation & Cart & Trolley \\
        \midrule
        Arm          & UR10 model  & UR10      & UR5 \\
        Base         & Rigid wheels & Aluminium & Plywood \\
        Support      & Rigid floor  & Concrete  & Foam \\
        Resistance   & Varied       & Low, high & Low, high \\
        Sander speed & Varied       & Low, high & High \\
        \bottomrule
    \end{tabular*}
    \caption{Simulation model and physical platforms.}
    \label{tab:platforms}
\end{table}

\subsection{Platforms}
\label{sec:platforms}

The cart carries a UR10 on an aluminium frame and deck with four lockable wheels, on a concrete floor (Figure~\ref{fig:physical_setup}a).

The trolley carries a UR5 on a lighter plywood deck with four lockable wheels (Figure~\ref{fig:physical_setup}b). A foam layer beneath the trolley provides a repeatable compliant support that stands in for compliant tyres, suspension, or soft ground, the non-rigid components motivating the base-compliance study of Wang and Carmichael \shortcite{Flexible_Experiment_Wang}.

\subsection{Experimental Setup}

Both platforms use a wall-mounted panel and the same wrist-mounted RYOBI ROS300-S random-orbital sander (300~W, 125~mm pad), operated at its lowest and highest dial settings, referred to as low and high tool speed.\footnote{Manufacturer's specification: no-load speeds of approximately 7{,}000 and 13{,}000~min$^{-1}$ and orbital stroke rates of 14{,}000 and 26{,}000~min$^{-1}$, respectively.} The base--ground resistance is set low by leaving the wheels free and high by locking them. Locking the wheels increases the resistance to base motion but does not rigidly fix the platform to the ground; compliance in the wheels, deck, and support remains.

\begin{figure*}[tp]
    \centering
    \begin{tabular}[b]{@{}c@{}}\includegraphics[height=4cm]{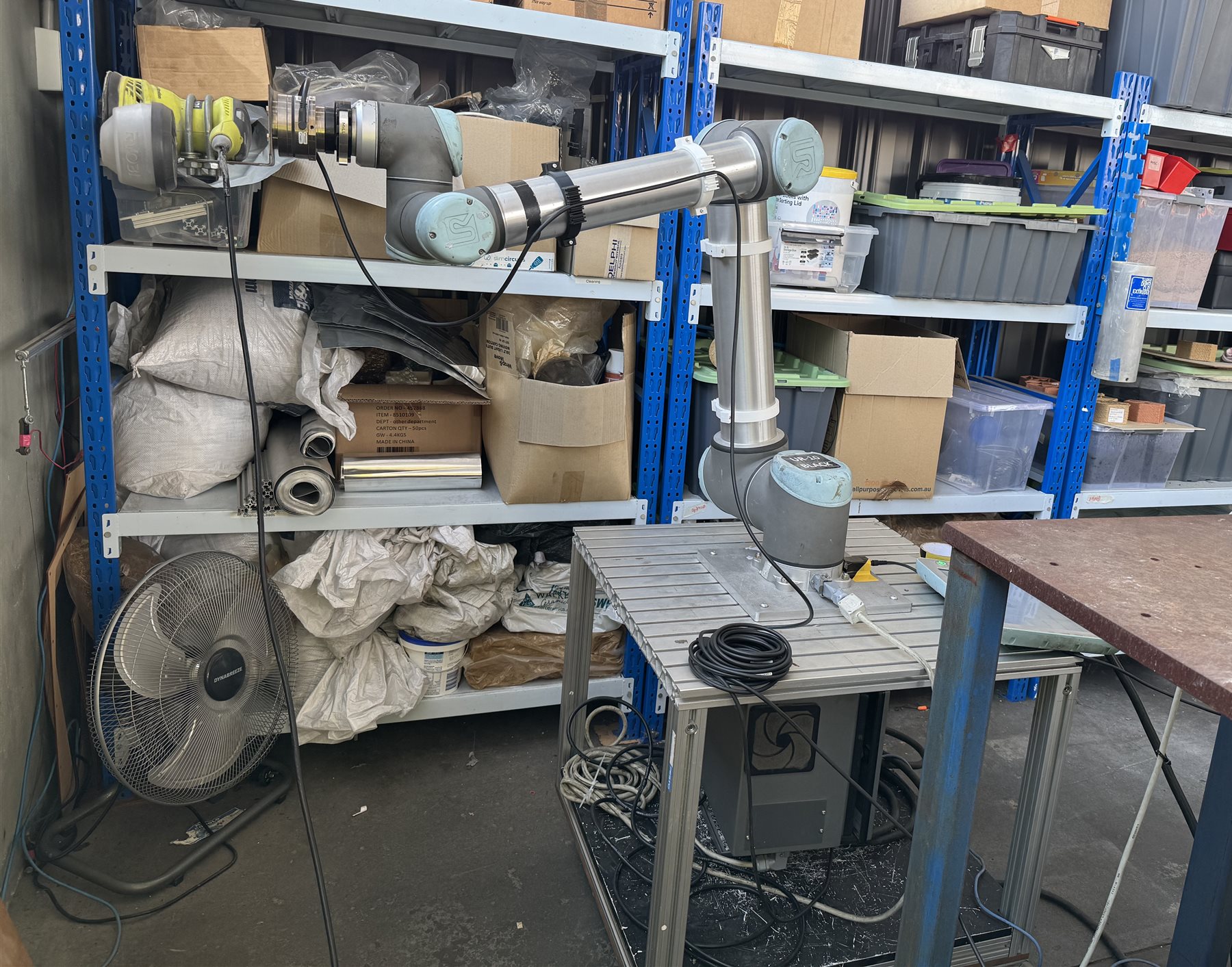}\\[1pt]\small (a) Cart (UR10)\end{tabular}\hfill
    \begin{tabular}[b]{@{}c@{}}\includegraphics[height=4cm]{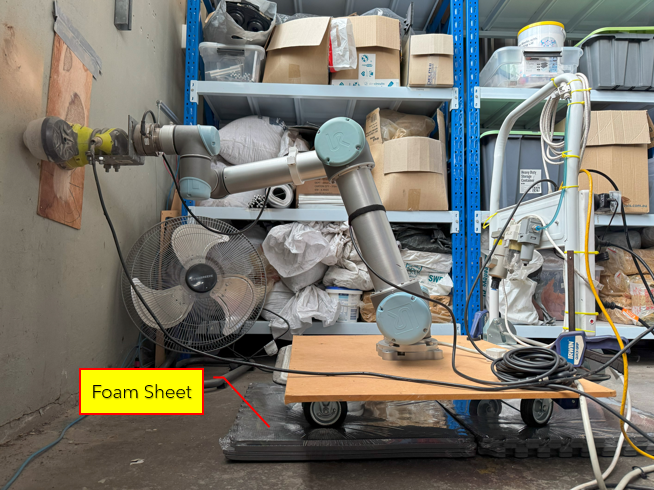}\\[1pt]\small (b) Trolley (UR5) on foam\end{tabular}\hfill
    \begin{tabular}[b]{@{}c@{}}\includegraphics[height=4cm]{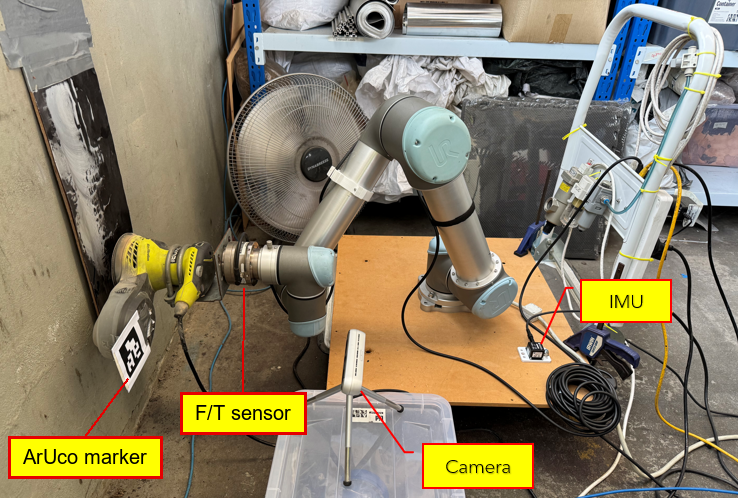}\\[1pt]\small (c) Sensors\end{tabular}
    \caption{Physical testbeds. The cart has an aluminium-extrusion frame with a slotted aluminium deck; the trolley has a plywood deck with the compliant layer beneath the base. Panel (c) shows the force/torque sensor, deck-mounted inertial measurement unit, tripod-mounted camera, and tool-mounted ArUco marker used with the trolley; the same sensors were deployed on the cart.}
    \label{fig:physical_setup}
\end{figure*}

A wrist-mounted Axia80-M20 six-axis force/torque sensor (ATI Industrial Automation; effective resolution 0.1~N and 0.005~N\,m) measures contact force. A deck-mounted Orientus inertial measurement unit (Advanced Navigation; $\pm$4\,$g$ range) records base acceleration at 1000~Hz, with no additional software filtering applied to the recorded data. An Intel RealSense D435 camera on a fixed tripod tracks a tool-mounted ArUco marker to measure tool motion (Figure~\ref{fig:physical_setup}c).

\subsection{Force Controller}

A task-space admittance controller converts normal-force error into end-effector velocity while maintaining tangential position. Cartesian commands are resolved to joint velocities using a quadratic programme \cite{Reactive_control1}.

\subsection{Task and Trials}
\label{sec:trials}

The controller regulates normal force against the wall while holding a fixed tangential target relative to the base. Normal end-effector motion is commanded relative to the base, so motion of the base relative to the wall changes the resulting tool--workpiece displacement. The TCP error is estimated from the tool-mounted marker, tracked by the fixed camera, as its displacement from a reference position (Section~\ref{sec:metrics}). It is therefore measured relative to the workpiece, whereas the tangential target is held relative to the base; base motion thus enters the TCP error even when the arm tracks its own command.

The trials are organised into three experiments, summarised in Table~\ref{tab:results}. Experiment~1, on the cart, varies the base--ground resistance, arm posture, and commanded force in single-condition trials (T1--T4). Experiment~2, also on the cart, switches from low to high resistance part-way through contact at two tool speeds (T5, T6), providing a within-trial comparison of the two resistance conditions while tracking TCP error with the camera. Experiment~3 examines whether the same effects are observed on a lighter and more compliant platform. The trolley is lighter than the cart and is placed on foam to introduce additional support compliance. T7 and T8 repeat the low- and high-resistance conditions at high tool speed. A separate hard-floor and foam comparison at low resistance, high tool speed, and a 20~N command examines the effect of support compliance.

\begin{table*}[!t]
    \centering
    \footnotesize
    \setlength{\tabcolsep}{3pt}
    \begin{tabular*}{\textwidth}{@{\extracolsep{\fill}}l l l c c c c c c c c c c@{}}
        \toprule
        & \multicolumn{4}{c}{Condition} & \multicolumn{8}{c}{Result} \\
        \cmidrule(r){2-5}\cmidrule(l){6-13}
        ID & Resistance & Posture & Speed & $F^{*}$ & $\bar{F}$ & $\rho_F$ & $e_F$ & $\sigma_F$ & Slope & $e_{\mathrm{TCP}}$ & \multicolumn{2}{c}{$a_{pp}$ (m\,s$^{-2}$)} \\
        \cmidrule(l){12-13}
        & & & & (N) & (N) & & (N) & (N) & (N\,s$^{-1}$) & (mm) & med. & max \\
        \midrule
        \multicolumn{13}{@{}l}{\emph{Experiment 1 (cart)}} \\
        T1 & Low  & Standard & Low & 100 & 58.8 & 0.59 & 41.5 & 2.6 & 0.42 & n/a & 0.19 & 0.69 \\
        T2 & Low  & Low      & Low & 100 & 68.3 & 0.68 & 32.1 & 3.7 & 0.32 & n/a & 0.08 & 0.12 \\
        T3 & High & Standard & Low & 100 & 94.3 & 0.94 & 6.9  & 3.2 & 0.21 & n/a & 0.29 & 0.77 \\
        T4 & Low  & Low      & Low & 25  & 26.3 & 1.05 & 1.9  & 1.3 & $-0.04$ & n/a & 0.06 & 0.11 \\
        \midrule
        \multicolumn{13}{@{}l}{\emph{Experiment 2 (cart)}} \\
        T5 & Low            & Low & Low & 100 & 58.5  & 0.58 & 41.6 & 3.3 & 0.06 & 98  & 0.10 & 0.17 \\
           & High from 45~s &     &   &     & 88.3  & 0.88 & 14.1 & 6.9 & 0.77 & 1   & 0.44 & 0.92 \\
        T6 & Low            & Low & High & 100 & 69.7  & 0.70 & 30.6 & 3.5 & 0.26 & 103 & 0.35 & 0.51 \\
           & High from 39~s &     &   &     & 100.8 & 1.01 & 7.7  & 6.1 & 0.75 & 1   & 3.9  & 5.9 \\
        \midrule
        \multicolumn{13}{@{}l}{\emph{Experiment 3 (trolley on foam)}} \\
        T7 & Low  & Low & High & 100 & 53.7 & 0.54 & 47.2 & 4.0 & 0.77 & 96 & 18.5 & 23.4 \\
        T8 & High & Low & High & 100 & 94.4 & 0.94 & 9.9  & 7.3 & 0.35 & 1  & 24.0 & 28.4 \\
        \bottomrule
    \end{tabular*}
    \caption{Trials by experiment, with force realisation, tool centre point (TCP) error, and base vibration during contact, using the metrics defined in Section~\ref{sec:metrics}.}
    \label{tab:results}
\end{table*}

Table~\ref{tab:results} lists the experimental conditions: low or high base--ground resistance, standard or low arm posture, two tool speeds, and force commands of 25~N or 100~N. The standard posture places the tool higher and farther from the base, while the low posture places it closer to the deck, reducing the moment arm of the contact force about the base. In each experiment, base vibration was also recorded with the sander running out of contact, as a reference for the base response to tool operation without workpiece contact. Each trial was performed once, so the results are treated as a case series with comparisons across conditions and experiments.

\begin{figure*}[tp]
    \centering
    \includegraphics[width=\textwidth]{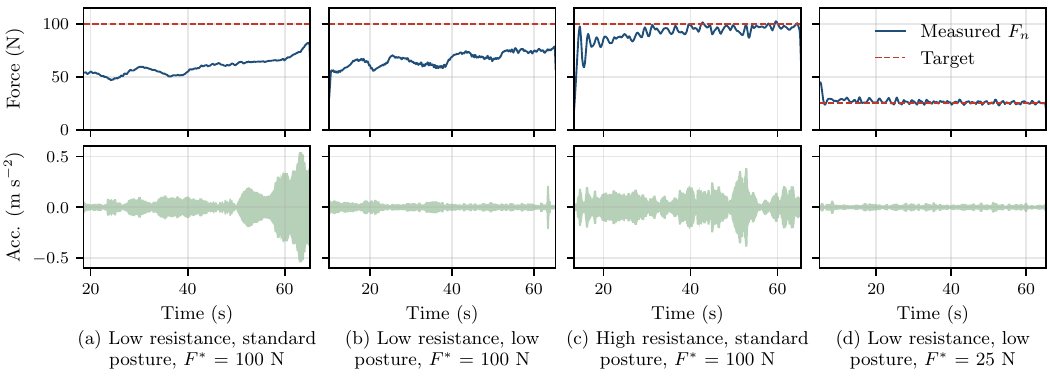}
    \caption{Experiment~1: single-condition trials, each shown during contact up to retraction. Top: measured normal force and command. Bottom: base acceleration envelope normal to the deck. In each row, (a)--(d) share the vertical axis.}
    \label{fig:base_conditions}
\end{figure*}

\subsection{Evaluation Metrics}
\label{sec:metrics}

The realised force ratio $\rho_F = \bar{F}/F^{*}$ is the mean normal force $\bar{F}$ divided by the commanded force $F^{*}$. Force error $e_F$ is the root-mean-square deviation from the command. Force variation $\sigma_F$ is the standard deviation of the force after subtracting a least-squares linear fit over the evaluation interval, while the slope of that fit is reported separately as force creep. All metrics are computed over steady contact, excluding the transients at contact onset, at the resistance switch, and at retraction; for T5 and T6, the low- and high-resistance phases are evaluated separately.

Base vibration $a_{pp}$ is the peak-to-peak acceleration normal to the deck over 1~s intervals, summarised by its median and maximum. Base vibration is compared within each experiment. TCP error $e_{\mathrm{TCP}}$ is the magnitude of the tool-marker displacement over the evaluation interval (Section~\ref{sec:trials}); for the high-resistance phases of T5 and T6 it is therefore the displacement added after the switch, whereas the figures show the displacement accumulated from the start of the record.

\section{Results}

The trials revealed persistent force shortfall and tool motion at low base--ground resistance, both of which changed when the resistance was increased during contact. Experiment~2 (Figure~\ref{fig:locking}) shows this sequence within single trials. Base vibration varied with the tool, support, and contact conditions.

\subsection{Force Shortfall and TCP Motion at Low Resistance}

At a 100~N command, all low-resistance trials realised only 54--70\% of the commanded force (Table~\ref{tab:results}). The shortfall persisted through steady contact. In T1, force rose to about 50~N and then crept upwards at 0.4~N\,s$^{-1}$, giving a realised force of 58.8~N and a force error of 41.5~N (Figure~\ref{fig:base_conditions}a). Force variation was 2.6~N, indicating that the principal departure from the command was a sustained force shortfall rather than large fluctuations around the target.

\begin{figure*}[t]
    \centering
    \includegraphics[width=\textwidth]{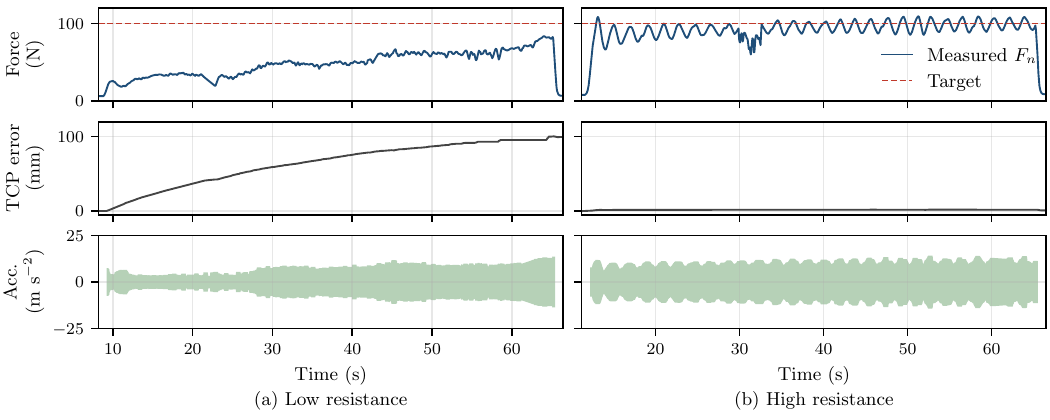}
    \caption{Experiment~3. Top: normal force. Middle: TCP error, the displacement of the tool marker from its position at the start of the record. Bottom: base acceleration envelope normal to the deck. In each row, (a) and (b) share the vertical axis.}
    \label{fig:session3}
\end{figure*}

On the second platform, T7 realised 54\% of the command and accumulated 96~mm of TCP error (Figure~\ref{fig:session3}a). The error initially grew at 2.9~mm\,s$^{-1}$ before slowing, while force continued to rise, with a force creep of 0.8~N\,s$^{-1}$.

\begin{figure*}[t]
    \centering
    \includegraphics[width=\textwidth]{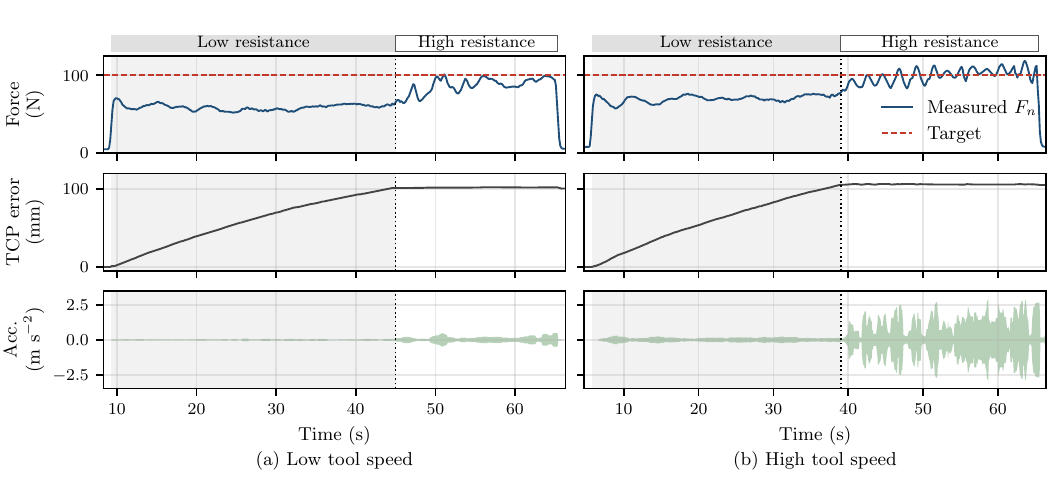}
    \caption{Experiment~2: switch from low to high resistance part-way through contact. The bar above each panel marks the low-resistance (shaded) and high-resistance phases; the dotted line marks the switch. Top: normal force. Middle: TCP error, the displacement of the tool marker from its position at the start of the record, tracked by the fixed camera. Bottom: base acceleration envelope normal to the deck. In each row, (a) and (b) share the vertical axis.}
    \label{fig:locking}
\end{figure*}

Experiment~2 shows the same combination within single trials (Figure~\ref{fig:locking}). In the low-resistance phases of T5 and T6, the realised force ratios were 0.58 and 0.70, while TCP error accumulated to 98 and 103~mm at 2.8--3.1~mm\,s$^{-1}$ with the tangential target fixed relative to the base. Sustained contact with relatively low short-term force variation, 3.3 and 3.5~N, thus coexisted with continued tool motion across the workpiece, until the resistance was increased part-way through contact.

\subsection{Response to Increased Base Resistance}
\label{sec:locked}

The within-trial switches in Experiment~2 provide the most direct comparison of the two resistance conditions. In T5 and T6, resistance was increased at 45~s and 39~s, respectively, while the force command remained at 100~N. TCP-error growth was arrested within about 1~s, with only about 1~mm of additional displacement over each high-resistance evaluation interval. Force rose towards the command over about 5~s, giving realised force ratios of 0.88 and 1.01.

Force error fell from 41.6 to 14.1~N in T5 and from 30.6 to 7.7~N in T6, reductions of 66\% and 75\%. Force variation increased from 3.3 to 6.9~N and from 3.5 to 6.1~N, respectively, and force creep was 0.75--0.77~N\,s$^{-1}$. The substantially lower force error indicates that overall force tracking improved at both tool speeds.

The single-condition comparisons showed the same direction of change. From T1 to T3, the realised force ratio increased from 0.59 to 0.94 and force error decreased from 41.5 to 6.9~N (Figure~\ref{fig:base_conditions}c). From T7 to T8, the realised force ratio increased from 0.54 to 0.94 and force error decreased from 47.2 to 9.9~N; T8 also showed about 1~mm of TCP error (Figure~\ref{fig:session3}b). The same direction of change was therefore observed in the within-trial switches, in the separate trials, and on both platforms.

\subsection{Dependence on Posture and Commanded Force}

Changing from the standard to the low posture reduced the force shortfall at low resistance. From T1 to T2, the realised force ratio rose from 0.59 to 0.68 and force error fell from 41.5 to 32.1~N (Figure~\ref{fig:base_conditions}a,b). Force still crept upwards at 0.3~N\,s$^{-1}$ in T2 and remained well below the 100~N command. The posture change altered the response without removing the sustained force shortfall.

The commanded-force comparison showed a larger change. T4 differed from T2 only in the commanded force, reduced from 100 to 25~N. At 25~N, the realised force was 26.3~N, force error was 1.9~N, and force variation was 1.3~N, with negligible force creep (Figure~\ref{fig:base_conditions}d). The low-resistance condition therefore produced substantially different force-tracking behaviour at the two loads: a persistent force shortfall at 100~N and close tracking at 25~N.

\begin{table*}[!t]
    \centering
    \footnotesize
    \setlength{\tabcolsep}{8pt}
    \renewcommand{\arraystretch}{1.1}
    \begin{tabularx}{\textwidth}{@{}>{\raggedright\arraybackslash}p{0.16\textwidth}>{\raggedright\arraybackslash}p{0.32\textwidth}>{\raggedright\arraybackslash}X@{}}
        \toprule
        Aspect & Simulation & Hardware observations \\
        \midrule
        Base motion & Slip when the required wheel--ground tangential forces exceed the available traction & About 100~mm TCP error at low resistance (T5--T7), accompanied by $\rho_F$ of 0.54--0.70 (T1, T2, T5--T7), on both platforms \\
        \rowsep
        Posture sensitivity & Extended posture increases excitation and reduces the stability margin & Low posture reduced median base vibration from 0.19 to 0.08~m\,s$^{-2}$ and raised $\rho_F$ from 0.59 to 0.68 (T1, T2) \\
        \rowsep
        Base--ground interaction & Higher wheel friction suppresses slip & High resistance raised $\rho_F$ to 0.88--1.01 on both platforms (T3, T5, T6, T8); the switch arrested TCP-error growth within about 1~s (T5, T6) \\
        \rowsep
        Oscillation / tool excitation & Tool excitation produced tool and base oscillation; tool speed shifted the dominant frequency & High tool speed produced substantially higher base vibration than low tool speed, both out of contact and at high resistance (T5, T6) \\
        \rowsep
        Support compliance & Not represented in the rigid-wheel model & Foam nearly tripled median base vibration relative to the hard floor (Figure~\ref{fig:compliance}) \\
        \rowsep
        Loss of stability & Occurred when the zero-moment point moved outside the support polygon & Not reached \\
        \bottomrule
    \end{tabularx}
    \caption{Qualitative comparison of simulated behaviours and hardware observations.}
    \label{tab:sim_vs_exp}
\end{table*}

\subsection{Vibration Across Tool and Support Conditions}
\label{sec:compliance}

Median base vibration increased in every low-to-high resistance comparison: from 0.19 to 0.29~m\,s$^{-2}$ in T1/T3, 0.10 to 0.44~m\,s$^{-2}$ in T5, 0.35 to 3.9~m\,s$^{-2}$ in T6, and 18.5 to 24.0~m\,s$^{-2}$ in T7/T8. These increases accompanied higher realised forces and lower force errors. The vibration also evolved during contact: T1 developed pronounced bursts near the end despite its lower median vibration, while in T7 vibration increased as the force rose and TCP motion slowed.

Tool speed produced a pronounced difference in Experiment~2. With the sander running out of contact, median base vibration was approximately 0.03~m\,s$^{-2}$ at low tool speed and 1.7~m\,s$^{-2}$ at high tool speed. During high-resistance contact, the corresponding values were 0.44 and 3.9~m\,s$^{-2}$. Despite this nearly ninefold difference, force variation was slightly lower at high tool speed, at 6.1~N compared with 6.9~N at low tool speed. Larger base vibration therefore did not correspond to larger force variation in this comparison.

\begin{figure}[!t]
    \centering
    \includegraphics[width=\columnwidth]{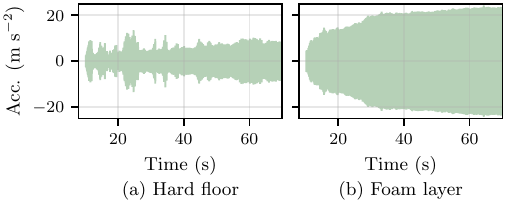}
    \caption{Experiment~3, support comparison (low resistance, 20~N command): base acceleration envelope normal to the deck.}
    \label{fig:compliance}
\end{figure}

Posture and support also changed the vibration response. Lowering the arm from T1 to T2 reduced median base vibration from 0.19 to 0.08~m\,s$^{-2}$ while the realised force increased. At the lower load in T4, median base vibration was 0.06~m\,s$^{-2}$, below the 0.13~m\,s$^{-2}$ measured with the sander running out of contact in Experiment~1. In the separate support comparison in Experiment~3, at low resistance and a 20~N command, median base vibration increased from 16~m\,s$^{-2}$ on the hard floor to 42~m\,s$^{-2}$ on foam (Figure~\ref{fig:compliance}).

Contact loading could also reduce vibration. In Experiment~3, median base vibration was approximately 45~m\,s$^{-2}$ with the sander running out of contact, falling to 18.5 and 24.0~m\,s$^{-2}$ during contact in T7 and T8. Together, these observations show a dynamic response that depends on tool excitation, support, and contact state, and does not vary consistently with force error or force variation.

\subsection{Relation to the Simulation}

Table~\ref{tab:sim_vs_exp} compares the simulated behaviours with the hardware observations. Both studies showed sensitivity to base--ground interaction and arm posture. Tool excitation produced oscillatory responses in simulation, while tool speed strongly affected base vibration in hardware. The hard-floor and foam comparison additionally exposed a dependence on support compliance, which the rigid-wheel model does not represent. Loss of stability was not observed in the physical trials. In simulation, this behaviour occurred under more adverse combinations of base and loading parameters than those used in the physical experiments. These conditions were not deliberately reproduced on hardware for safety reasons.

\section{Discussion}

The central observation is that sustained contact did not guarantee that the commanded force was realised or that the intended tool position relative to the workpiece was maintained. Under the 100~N command at low resistance, contact was maintained throughout while the realised force remained well below the command and the tool moved about 100~mm across the workpiece. The within-trial resistance changes provide the clearest link between base--ground resistance and contact execution: with the force command unchanged, TCP-error growth was arrested and realised force increased. The same direction of change was observed on the second platform.

The simultaneous force shortfall and TCP motion, and their response to increased resistance, are consistent with base motion absorbing part of the commanded tool advance, as outlined in the Introduction. Lowering the arm changes the contact-force moment about the wheel contacts, while reducing the commanded load changes the reaction that the support must sustain. The posture and load comparisons show how these task conditions affect the expression of non-rigid base effects.

Vibration characterises another part of the response. Tool excitation is evident even out of contact, and the hard-floor and foam comparison shows the influence of support conditions. The reduction in vibration when the pad was loaded in Experiment~3 is consistent with contact changing how excitation is transmitted through the tool, arm, base, and support. The T1 end-of-contact bursts also show that vibration can evolve substantially within one resistance condition. The improved force tracking at high resistance, despite increased vibration, illustrates how these measurements describe different aspects of contact.

The three measurements consequently serve complementary purposes. Force quantifies how closely the contact load follows the command; the tool marker reveals continued tool motion relative to the workpiece even when the contact force appears to have plateaued; and base vibration describes the response to tool excitation and support conditions. Their joint measurement makes the consequences of non-rigid base behaviour visible beyond a single tracking metric. The simulation supplies a qualitative framework for base motion, oscillation, and loss of stability, while the hardware comparisons reveal compliance-related responses that are not represented by the rigid-wheel model.

The scope of these conclusions is bounded by the case-series design. Each condition was run once, with the principal force and motion responses observed on two platforms. The measured tool motion includes contributions from manipulator tracking error, arm deflection, and base motion. Realised force and arm configuration also changed across the resistance conditions, so their contributions to the observed vibration cannot be separated from the effect of increased base--ground resistance. All contact trials used the operating sander, leaving tool and controller contributions to force variation unresolved. Surface finish and material removal were not assessed, so vibration amplitude is not established as a measure of process quality. The two platforms and support conditions cover only a limited range of lightweight mobile manipulator configurations.

\section{Conclusion and Future Work}
\label{sec:conclusion}

This sanding case study characterised non-rigid base effects during contact tasks on two lightweight wheeled manipulation testbeds. At low base--ground resistance, a 100~N command produced persistent force shortfall, with only 54--70\% realised and TCP errors of about 100~mm in the camera-tracked trials. In the within-trial comparisons, increasing resistance during contact arrested further TCP-error growth within about 1~s and raised the average realised force to 88--101\% of the command during the high-resistance phases. Separate low- and high-resistance trials on the second platform showed the same direction of change. The magnitude of the force shortfall depended on posture and commanded force: the low posture reduced it, while a 25~N command was closely tracked at low resistance. Base vibration showed a distinct dependence on tool excitation, support, and contact loading. The exploratory simulation provided a qualitative framework for base motion, oscillation, and stability, while the hardware exposed compliance-related responses not represented by the rigid-wheel model. Together, these results characterise non-rigid base effects through force realisation, tool motion, and dynamic response.

Future work should examine commanded force, posture, and support conditions independently, repeat the comparisons across a wider range of platforms, and relate the measured dynamic response to surface finish and material removal. Experiments on actuated mobile robots should also evaluate coordinated wheel-motor and arm commands to compensate for measured base motion while maintaining contact force. Their effectiveness would need to be assessed against available wheel traction and base stability. These studies would help define feasible contact conditions for mobile manipulators operating with non-rigid base--environment support.

\section*{Acknowledgments}
This work was supported by the Australian Research Council under Grant No.\ IC200100001. The authors would like to acknowledge the support of Queensland University of Technology (QUT), the Australian Cobotics Centre (ACC), and the QUT Centre for Robotics (QCR). The authors further thank the Advanced Robotics for Manufacturing Hub (ARM Hub) for providing access to the experimental facilities.

\bibliographystyle{named-initials}
\bibliography{NRF}

\end{document}